\documentclass{article}
\usepackage[preprint]{neurips_2026}
\usepackage[utf8]{inputenc}
\usepackage[T1]{fontenc}
\usepackage{hyperref}
\usepackage{url}
\usepackage{booktabs}
\usepackage{amsmath,amssymb}
\usepackage{graphicx}
\usepackage[table]{xcolor}
\usepackage{multirow}
\usepackage{subcaption}
\usepackage{xspace}
\usepackage{wrapfig}
\usepackage{placeins}
\usepackage[most]{tcolorbox}
\usepackage{caption}
\DeclareCaptionType{prompt}[Prompt][List of Prompts]
\usepackage[resetlabels,labeled]{multibib}
\newcites{S}{Appendix References}

\newcommand{\initiating}{\textsc{Initiating}\xspace}
\newcommand{\qualifying}{\textsc{Qualifying}\xspace}
\newcommand{\grounding}{\textsc{Grounding}\xspace}
\newcommand{\inferring}{\textsc{Inferring}\xspace}
\newcommand{\hypothesizing}{\textsc{Hypothesizing}\xspace}
\newcommand{\backtracking}{\textsc{Backtracking}\xspace}
\newcommand{\constraining}{\textsc{Constraining}\xspace}

\newcommand{\xhdr}[1]{\vspace{-1mm}\noindent{{\bf #1.}}}

\definecolor{lightpurple}{RGB}{230,220,245}
\title{ReasonOps: Operator Segmentation for LLM Reasoning Traces}

\author{%
Daniel Lee \\
Stanford University \\
\texttt{leedan@stanford.edu}
\And
Owen Queen \\
Stanford University \\
\texttt{oqueen@stanford.edu}
\And
James Zou \\
Stanford University \\
\texttt{jamesz@stanford.edu}
}

\begin{document}

\maketitle

\begin{abstract}
Chain-of-thought traces from large reasoning models can span tens of thousands of tokens, yet we lack a vocabulary for describing their internal structure.
Previous methods developed to analyze chain-of-thought traces are either too rigid or not expressive enough, failing to capture features across domains and models.
To remedy this, we develop ReasonOps, an unsupervised, expressive method for annotating chain-of-thought traces, providing succinct universal operators.
Using ReasonOps, we analyze 44{,}662 traces from 12 thinking LLMs spanning 6 families across 8 reasoning benchmarks and discover that they share a common compositional structure: 7 recurring reasoning operators---discourse-level moves such as \backtracking, \inferring, and \hypothesizing---that emerge from unsupervised clustering of sentence-initial 3-token pivots.
These operators appear across every model family and benchmark domain, confirmed by three independent LLM judges who classify held-out samples at 70--76\% accuracy.
We analyze the structure of operators on easy vs. hard problems, revealing that reflective operators are more helpful on hard problems and harm performance on easy problems.
Reasoning traces are highly model-identifying: structural operator features plus anchor-phrase text features recover the source model with macro-AUC $= 0.987$, revealing that each model family has a distinctive reasoning fingerprint.
Structural operator features predict within-problem answer correctness well above baselines.
Classifiers built on these operators reach WP-AUC $= 0.701$ globally and $0.801$ on AIME.
ReasonOps further enables early quality estimation well before the trace completes: we predict at WP-AUC $= 0.664$ for only 50\% of the trace.
The ReasonOps pipeline is unsupervised and annotation-free, enabling deep insights into LLM reasoning traces as well as strong downstream results on model identification and correctness prediction.
\end{abstract}

\vspace{-2mm}
\section{Introduction}
\label{sec:intro}
\vspace{-2mm}

Reasoning-capable large language models (LLMs) increasingly solve difficult problems by producing long intermediate traces before emitting a final answer.
LLMs have become synonymous with "large reasoning models" (LRMs) as the majority of frontier models today are post-trained to elicit extended chain-of-thought reasoning before producing a final answer \cite{yang2025qwen3, team2026kimi, guo2025deepseek, agarwal2025gpt}.
Prompting methods, multi-path decoding, process supervision, and reinforcement-learning-based post-training have all improved performance on mathematics, science, and coding benchmarks, but they have also made reasoning traces longer, more expensive, and harder to analyze \cite{wei2022chain, lightman2023let, guo2025deepseek}.
This comes at an increasing demand for monitoring and oversight of LLM decision-making processes \cite{guan2025monitoring, korbak2025chain, bowman2022measuring}.

Since the advent of chain-of-thought (CoT) prompting \cite{wei2022chain}, reasoning traces have been hypothesized as a window into the problem-solving capabilities of LLMs.
Reasoning traces, or chain-of-thought traces, are valuable artifacts for black-box understanding of LLMs as they can be obtained without access to model weights.
These traces have been shown to contain rich information \cite{shojaee2025illusion, kim2026reasoning}, yet we lack a uniform vocabulary under which to characterize reasoning traces across models, domains, and datasets.
While some methods have attempted to provide systematic annotations for reasoning traces, these methods often rely on \textit{a priori} vocabularies and syntactic structures that are overfit to current models or domains \cite{yang2025demystifying, venhoff2025understanding, lee2025reasoningflow, bogdan2025thought, li2025understanding}.
In this paper, we ask: can we compress diverse traces into a small vocabulary of reasoning operators shared across model families, tasks, and domains while preserving predictive information about success and failure?

\xhdr{Our contributions} We introduce ReasonOps, an unsupervised framework for inducing a compact vocabulary of reasoning operators from visible chain-of-thought traces. We use sentence-initial 3-token pivots, frequency filtering, and semantic clustering with e5-small \citep{wang2024e5}. We show that the resulting 7 operators are quantitatively meaningful: they generalize across 12 thinking LLMs from 6 families and 8 benchmarks, confirmed by three independent LLM judges (70--76\% classification accuracy; chance: 14\%). Structural operator features plus anchor-phrase text features identify the source model with macro-AUC $= 0.987$, revealing model-family reasoning fingerprints.
We show that operator features predict within-problem correctness above all span-free baselines and the LLM self-judge (the OST reaches $0.701$ cross-dataset, matching the content-augmented Op-XGB classifier while reading only operator labels), and that the OST---trained once on full sequences---enables correctness estimation from partial traces at any depth, surpassing an Op-XGB upper bound retrained per depth. We open-source our codebase as a resource for the community.\footnote{Code: \url{https://github.com/lee-dan/ReasonOps}}

\begin{figure}[t]
\centering
\includegraphics[width=\linewidth]{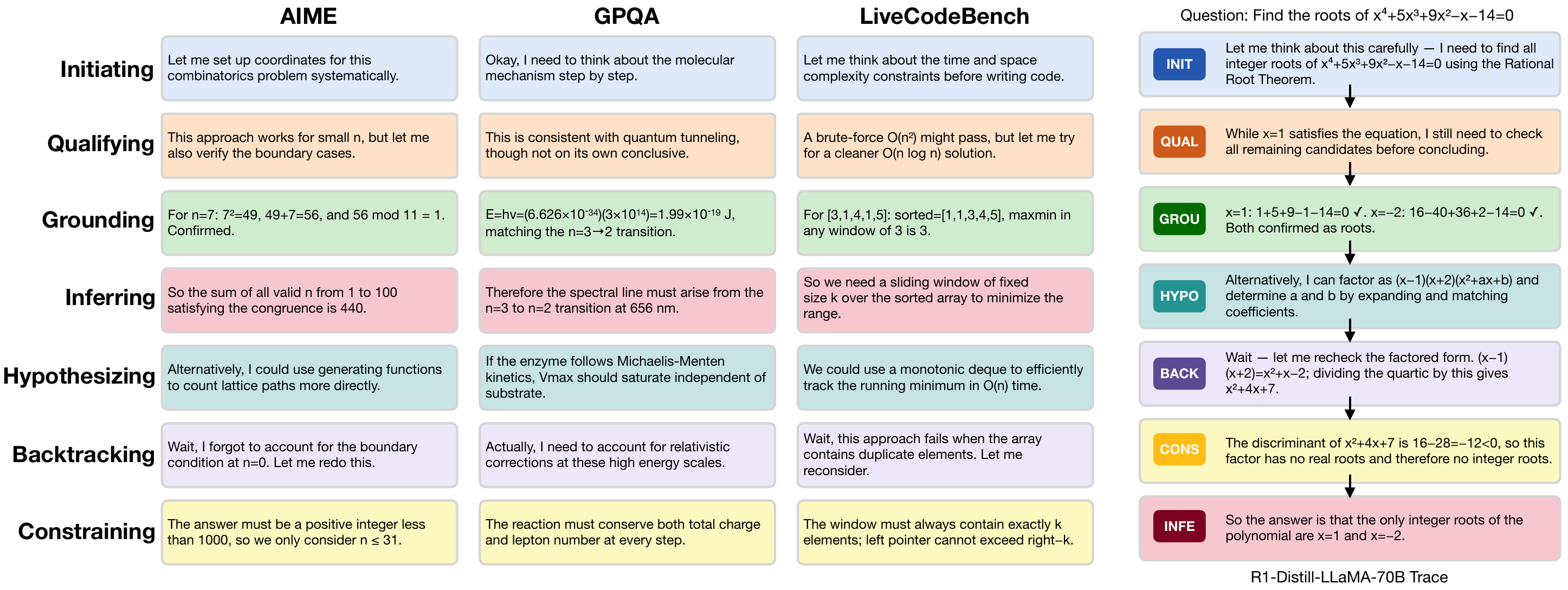}
\caption{(Left) representative operator units extracted from AIME, GPQA, and LiveCodeBench. (Right) representative reasoning trace extracted from R1-Distill-LLaMA-70B on MATH dataset.}
\label{fig:highlight}
\end{figure}

\vspace{-2mm}

\FloatBarrier
\section{Related Work}

\vspace{-2mm}

\xhdr{Annotating and analyzing reasoning traces}
Deepseek-R1 creators noted that "Wait" phrases were emergent behavior associated with reasoning capability increases \cite{guo2025deepseek}.
This and the interpretability of chain-of-thought traces inspired work into annotating and describing reasoning traces.
However, previous works primarily rely on \textit{a priori} vocabularies of reasoning \cite{yang2025demystifying, venhoff2025understanding, kargupta2025cognitive}, operate on arbitrary syntactic features such as sentences \cite{lee2025reasoningflow, bogdan2025thought}, or rely on domain-specific vocabularies \cite{li2025understanding}.
ReasonOps is an unsupervised method that does not assume \textit{a priori} syntactic features or a fixed vocabulary of annotations.
Other works have included \textit{ad hoc} methods to analyze reasoning traces for particular hypotheses \cite{shojaee2025illusion, kim2026reasoning}.

\xhdr{Monitoring and scalable oversight} The scalable oversight community has extensively explored ways to characterize LLM behavior \cite{bowman2022measuring, kenton2024scalable, engels2026scaling}, and chain-of-thought monitoring has been explored as a prominent direction \cite{korbak2025chain, guan2025monitoring}. This work is tangential to ReasonOps in terms of reasoning trace characterization.
Some work has also shown that chain-of-thought can be an unfaithful representation of model behavior \cite{lanham2023measuring}, but recent work has attempted to improve this in frontier models \cite{guan2025monitoring, paul2024making}.
Our work occupies a complementary point: we treat visible traces as behaviorally meaningful artifacts and ask whether they admit a stable, annotation-free meso-scale abstraction.

\xhdr{Correctness prediction} Some methods have been proposed for early correctness prediction, including LLM-only methods \cite{miao2024selfcheck, xiang2026thinking}, and mechanistic interpretability techniques \cite{zhao2026verifying}.
Similar to this line of work is that of process reward models \cite{lightman2023let} and outcome reward models \cite{cobbe2021training}, including work on verifiers in test-time scaling \cite{zhang2025generative, hosseini2024vstar}.
This work has been particularly prolific in mathematics problems \cite{cobbe2021training, lightman2023let, wang2024math}.
Other work has explored test time scaling techniques for self-verification, including SelfCheck \cite{miao2024selfcheck} and others \cite{wang2023self, brown2024large, snell2024scaling}.

\vspace{-2mm}

\FloatBarrier
\section{Methods}
\label{sec:method}

\vspace{-2mm}

We now describe how we build an unsupervised mechanism to infer operators from reasoning traces across 12 LLMs. Our pipeline is driven by linguistic principles and requires no \textit{a priori} definition of an operator vocabulary.

\xhdr{Data collection} We collect reasoning traces from 12 thinking language models spanning 6 families (Table~\ref{tab:models} in Appendix~\ref{app:models}) on 8 benchmarks: MATH-500 \citep{hendrycks2021math}, GPQA Diamond \citep{rein2024gpqa}, AIME 2024, LiveCodeBench \citep{jain2024livecodebench}, HumanEval \cite{chen2021evaluating}, MMLU-Pro \cite{wang2024mmlu}, ARC-Challenge \citep{clark2018arc}, and BIG-Bench Hard \cite{suzgun2023challenging}. All models are queried with a 65{,}536-token context budget. Raw chain-of-thought tokens are collected via the OpenRouter API for all models except Claude, which is queried directly via the Anthropic API with extended thinking enabled; per-model API details are given in Appendix~\ref{app:models}. After removing truncated and malformed traces, we retain \textbf{44{,}662 traces} ($\approx$3{,}720 per model).

\xhdr{Pivot-based span segmentation} Each trace is segmented into \textbf{spans}. A span begins at a \emph{pivot sentence}: any sentence whose first three alpha tokens (the \emph{sentence start}) appear frequently enough across the corpus to be recognized as a discourse signal. Formally, $\mathrm{pivot}(s) = (w_1, w_2, w_3)$ where $w_i$ are the ordered lowercase alphabetic tokens at the start of sentence $s$. A discourse pivot $p$ is included if (1) it appears in $\geq 100$ distinct traces (\textit{frequency filter}), (2) it appears in traces from $\geq 3$ distinct datasets (\textit{domain-diversity filter}), and (3) all three tokens $w_i$ belong to the top-2{,}000 most frequent corpus tokens (\textit{vocabulary filter}).

These thresholds are set to balance specificity and generality. The frequency threshold of 100 traces is chosen to exclude idiosyncratic phrases that appear in only a handful of traces while retaining common discourse moves; lowering it to 50 adds many domain-specific phrases that are not discourse-functional. The domain-diversity filter of 3 datasets ensures that any accepted pivot spans multiple task types, preventing dataset-specific phrases from entering the vocabulary. The top-2{,}000 vocabulary filter ensures pivots are composed of common English function words rather than domain-specific content tokens (e.g., chemistry formulae, programming keywords); we verified that varying this between 1{,}500 and 3{,}000 does not qualitatively change the resulting clusters.

\xhdr{Semantic pivot embedding and clustering} We embed all discourse pivots using the \texttt{intfloat/e5-small-v2} sentence encoder \citep{wang2024e5}.
Sentence-level embeddings capture full-phrase meaning, correctly separating pivots such as ``let me think'' and ``let me verify'' that share a common prefix. We cluster pivot embeddings with $K$-means ($K \in \{6, \ldots, 11\}$, 30 restarts), selecting $K$ by maximizing Cohen's $\kappa$ against an independent LLM judge on held-out spans. We select $K$ using Cohen's $\kappa$ with Claude Sonnet~4.6 as judge; $K{=}7$ maximizes $\kappa = 0.693$ (Appendix~\ref{app:ksweep}).\looseness=-1

\xhdr{Independence from correctness} The operator discovery pipeline uses no correctness labels: pivots are filtered by frequency and domain diversity, embeddings are fit on pivot text alone, and cluster assignments are by nearest-centroid lookup. Correctness labels enter only in the downstream prediction step (\S\ref{sec:downstream}), where Op-XGB and the OST are trained in a standard supervised 5-fold CV protocol. The operators themselves are thus a fully unsupervised intermediate representation.

\xhdr{Computational cost} The discovery pipeline is lightweight. Pivot extraction (frequency counting over all sentence starts) takes under 3 minutes on a single CPU core for 44{,}662 traces. Embedding the 5{,}464 accepted pivots with \texttt{e5-small-v2} takes 6 seconds; K-means on 5{,}464 points takes 12 seconds. Annotating a new trace requires only a dictionary lookup per span (no embedding at inference time), adding $<$1\,ms per trace. End-to-end operator discovery over the full corpus takes under 5 minutes, and annotating the entire 44K-trace corpus takes under 2 minutes.

\vspace{-2mm}

\section{Reasoning operators}
\label{sec:operators}

\vspace{-2mm}

\begin{table}[t]
\centering
\caption{The 7 discovered reasoning operators. Representative pivots are the most frequent 3-tuples in each cluster.}
\label{tab:operators}
\small
\setlength{\tabcolsep}{5pt}
\begin{tabular}{l>{\raggedright\arraybackslash}p{4.7cm}p{5.9cm}}
\toprule
Operator & Representative Pivots & Description \\
\midrule
\textsc{Initiating}    & let me think, let me check, let me verify & Explicitly launches a new cognitive operation \\
\textsc{Qualifying}    & hmm, but the problem, but let me & Introduces caveats or complications \\
\textsc{Grounding}     & i need to, the question is, looking at & Anchors reasoning in facts or given information \\
\textsc{Inferring}     & so the answer, thus final answer, yes & Draws a conclusion from prior steps \\
\textsc{Hypothesizing} & alternatively perhaps, for example if, so perhaps & Entertains a tentative or conditional scenario \\
\textsc{Backtracking}  & wait, wait no, wait let me & Signals potential error and prepares to restart \\
\textsc{Constraining}  & \mbox{we need to,} \mbox{now we need,} \mbox{so we need} & Identifies necessary conditions, narrows solution space \\
\bottomrule
\end{tabular}
\end{table}

Seven operators emerge from unsupervised clustering (Table~\ref{tab:operators}; Figure~\ref{fig:highlight}). They span the space of discourse moves in extended reasoning: goal-setting, fact-anchoring, inference-drawing, exploration, hedging, self-correction, and constraint-specification.
As a more coarse classification, operators can be broken into \textit{committal operators} (\initiating, \inferring, \constraining, \grounding) and \textit{reflective operators} (\qualifying, \backtracking, \hypothesizing).
These operators recur across all 12 models and all 8 benchmarks, suggesting that extended reasoning---regardless of training data, architecture, or problem domain---organizes itself into a common compositional vocabulary.
Figure~\ref{fig:highlight} shows representative span excerpts per operator across three of the eight benchmarks.
An illustration of a full operator sequence in Figure~\ref{fig:highlight} and Appendix~\ref{sec:appendix}.

\xhdr{Evaluation with an LLM judge} We first test whether the discovered operators are semantically coherent from the perspective of independent judges. Each LLM judge receives only cluster names, descriptions, and the top 12 representative pivot phrases per cluster---no training spans. It labels 50 held-out spans per cluster ($n = 350$ total).
Table~\ref{tab:kappa} reports classification accuracy for three judges. All reach 70--76\% on the 7-way classification task, 4.9--5.3$\times$ above the 14.3\% chance baseline.
Adding 3 example spans per cluster to the prompt raises accuracy from 69.8\% to \textbf{74.6\%}: few-shot examples aid pattern-matching from a closed 7-label set.

\begin{wraptable}{R}{0.5\textwidth}
\centering
\caption{LLM-judge validation ($n{=}350$ spans, 50 per cluster). Judges receive cluster names, descriptions, and top-12 representative pivots. \textbf{Exemplars}: prompt includes 3 example spans per cluster. The $K$-sweep is in Appendix~\ref{app:ksweep}.}
\label{tab:kappa}
\small
\begin{tabular}{llr}
\toprule
Judge & Prompt variant & Accuracy \\
\midrule
Random baseline   & ---        & 14.3\% \\
Claude Haiku~4.5  & baseline   & 76.0\% \\
GPT-5.4-Mini      & baseline   & 71.7\% \\
Claude Sonnet~4.6 & baseline   & 69.7\% \\
Claude Sonnet~4.6 & exemplars  & \textbf{74.6\%} \\
\bottomrule
\end{tabular}
\end{wraptable}

\xhdr{Cluster naming stability} To assess the semantic stability of the clusters, we re-ran the LLM naming step for the clusters by sampling 30 seeds of exemplars.
For each name and resulting description, we computed pairwise cosine similarity of description embeddings across the seeds.
The resulting descriptions were highly consistent within clusters with a median pairwise cosine similarity of $0.660$, substantially above both an across-cluster baseline of $0.467$ ($r_\mathrm{rb} = +0.72$) and a matched random-group control of $0.565$ ($r_\mathrm{rb} = +0.39$; $p < 10^{-149}$).
All seven operators showed significant within-cluster stability, indicating that the naming procedure recovers coherent and reproducible operator semantics; by contrast, randomly grouped spans did not elicit comparably coherent names.
See Appendix~\ref{app:stability} for more details.

\vspace{-2mm}

\section{Analysis of operator distributions}
\label{sec:analysis}
\vspace{-2mm}

\begin{figure}
    \centering
    \includegraphics[width=1.0\linewidth]{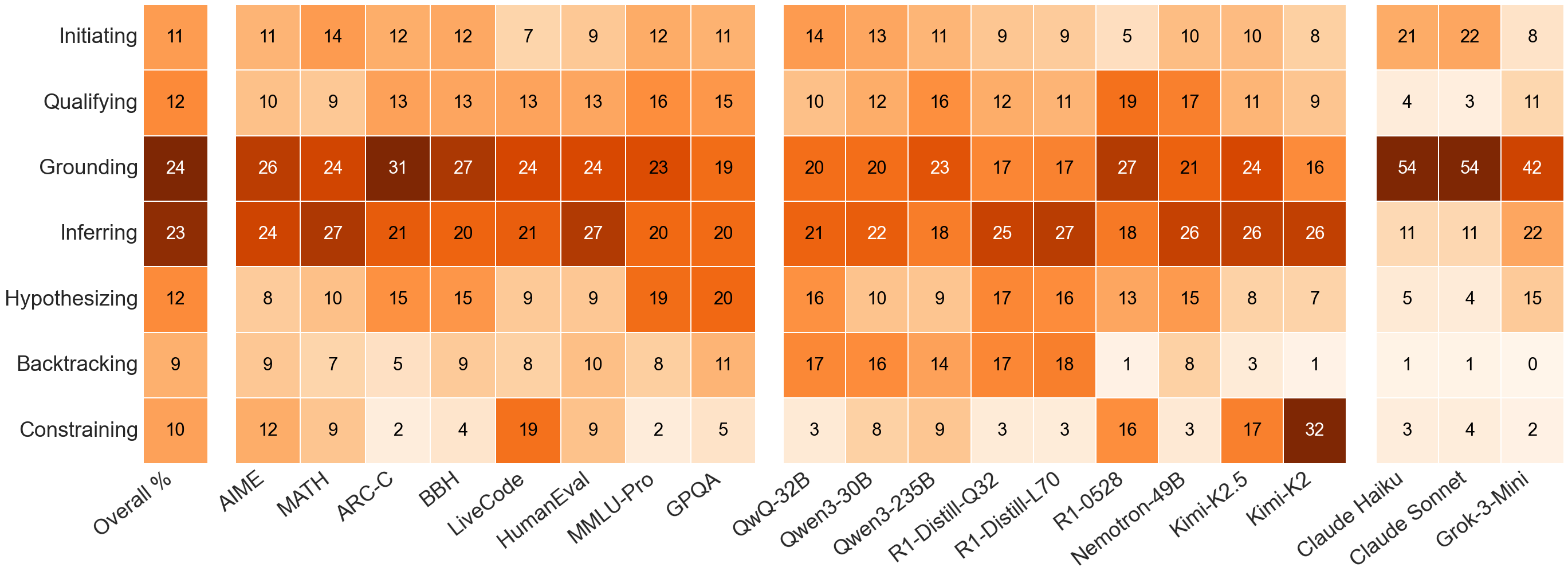}
    \caption{Distribution of operators across datasets (top) and models (bottom). Metrics are percent usage = average appearance of that operator for that given model/dataset vs. all other operators.}
    \label{fig:op_distribution}
\end{figure}

Operators show distinct patterns when stratified by dataset and model (Figure~\ref{fig:op_distribution}). Mathematical reasoning benchmarks (AIME, MATH) show higher \grounding and \constraining (step-by-step derivation), while commonsense benchmarks (ARC-C, BBH) use more \grounding and elevated \hypothesizing. Challenging multiple choice datasets like GPQA and MMLU-Pro exhibit higher use of \hypothesizing and \qualifying than any other dataset. Coding benchmarks (LiveCodeBench, HumanEval) exhibit elevated \constraining, consistent with specification-first programming. Models further show operator usage fingerprints that roughly cluster with model families, a phenomenon that's supported by the results in \S\ref{sec:model_id}.
R1-distill model profiles are both very similar, and Qwen models show almost identical operator proportions.
Kimi/Moonshot models show high \constraining usage.
Claude Haiku, Claude Sonnet, and Grok-3-Mini use excessive \grounding.
Despite these differences, all 7 operators are represented in every model.

\xhdr{Transition structure} The operator transition matrix reveals Markov structure in reasoning: \grounding strongly self-transitions (sustained fact-anchoring), while \backtracking most often leads into \initiating (correction followed by a fresh attempt). \hypothesizing frequently precedes \inferring, consistent with a generate-and-test schema. Run-length analysis reinforces this: \grounding and \constraining accumulate long self-runs, while \backtracking is almost always a single span---a brief interruption rather than a sustained state (Figure~\ref{fig:transition_rle}, Appendix~\ref{app:additional}). We expand on this \backtracking phenomenon in \S\ref{sec:backtracking}.

\begin{wrapfigure}{R}{0.5\textwidth}
    \centering
    \includegraphics[width=0.48\textwidth]{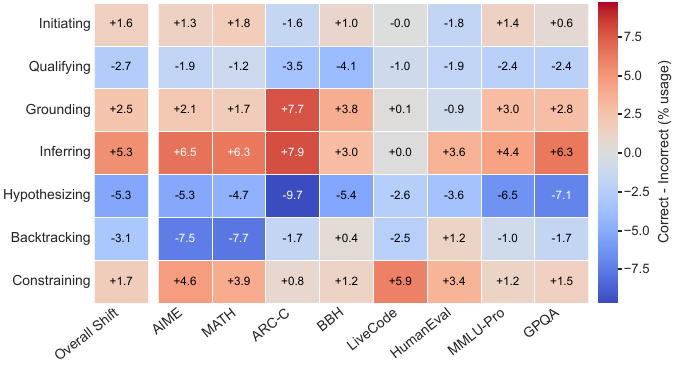}
    \caption{Shift of operator usage by benchmark in correct and incorrect traces. \% usage difference = take percent usage (Figure~\ref{fig:op_distribution}) and subtract usage in correct and incorrect traces.}
    \label{fig:op_heatmap_correctness}
\end{wrapfigure}

\vspace{-2mm}

\subsection{Operator usage on correct vs. incorrect traces}
\label{sec:temporal}

We see a few trends emerge in terms of operator usage on correct vs. incorrect traces (Figure \ref{fig:op_heatmap_correctness}). First, committal operators seem to be overrepresented in correct traces than in incorrect traces, with \grounding, \inferring, \initiating, and \constraining all being positive in correct - incorrect usage.
Conversely, reflective operators are globally seen more often in incorrect traces than correct traces. \hypothesizing and \qualifying show particularly higher usage in incorrect traces than correct traces.
This may be indicative of correct traces showing more confident, forward-stepping operations rather than reflecting.
This then led us to investigate these dynamics on easy vs. hard problems.

\begin{figure}
    \centering
    \includegraphics[width=1.0\linewidth]{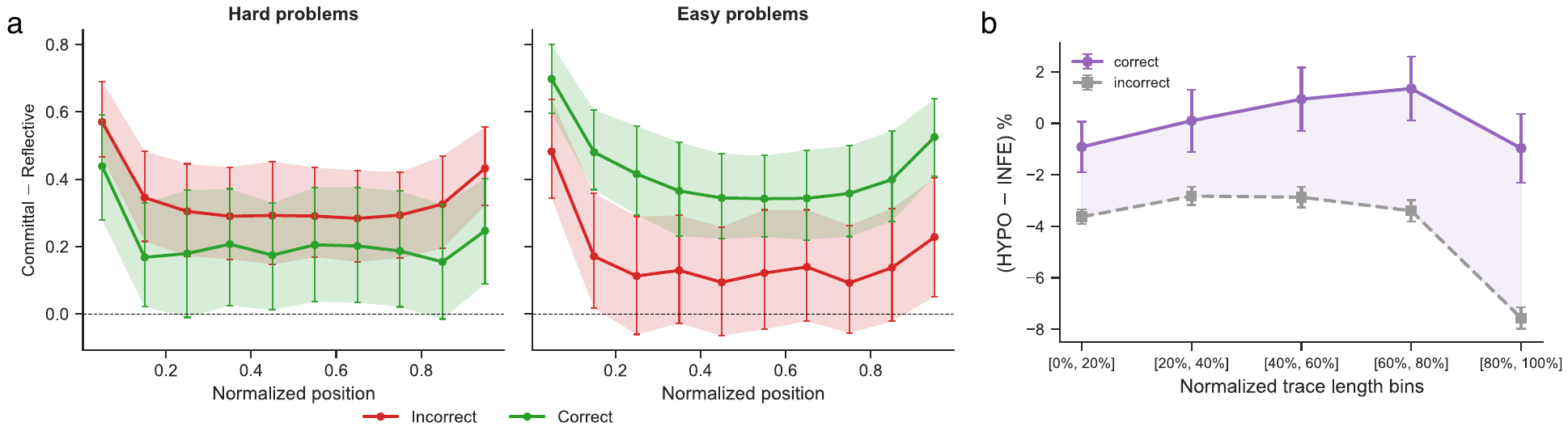}
    \caption{(a) Temporal distribution of committal - reflective operators for correct and incorrect problems. 95\% CI shown at each normalized position. (b) \hypothesizing - \inferring gap over only hard problems for correct and incorrect traces. Shaded regions show the gap between correct and incorrect curves.}
    \label{fig:hard_v_easy}
\end{figure}

\xhdr{Easy vs. hard problems} We next define a problem as ``hard'' if fewer than 1/3 of models solve it, and ``easy'' if more than 2/3 models pass.
We first filter to only BBH, LiveCodeBench, MMLU-Pro, and GPQA, which have sufficient samples of hard problems.
For easy problems, correct traces exhibit significantly higher usage of committal operators than reflective operators, indicating confident forward-stepping through the problem.
The usage of committal - reflective operators can be defined as the \textit{committal-reflective gap}.
On easy problems, this gap is much higher in correct traces ($+44.2\%$) than in incorrect traces ($+7.5\%$) and is strongly predictive of correctness (AUC = $0.76$ $[0.74, 0.78]$ 95\% CI). On average across all problems, committal usage peaks early and late in traces while reflective operators appear mid-trace (Figure~\ref{fig:hard_v_easy}a).

Hard problems show less obvious differences between correct and incorrect traces.
The committal-reflective gap flips, showing $+30.4\%$ on correct traces and $+34.9\%$ on incorrect traces, a $-4.5\%$ shift which is not statistically-significant (Mann-Whitney p=0.15) (Figure~\ref{fig:hard_v_easy}a).
The predictive signal disappears as the AUC for correct vs. incorrect via committal-reflective gap alone drops to AUC = $0.48$ $[0.44, 0.51]$.
However, examining \hypothesizing vs. \inferring, instead of all committal and reflective operators, yields some effects.
When examining the distribution of these operators over time, the \hypothesizing - \inferring gap is particularly representative for correct vs. incorrect hard problems.
This gap increases monotonically over the length of the trace, reaching $6.6\%$ at the tail of the trace, with incorrect traces showing much higher usage of \inferring over \hypothesizing (Figure~\ref{fig:hard_v_easy}b).
This shows that reflective operators, specifically \hypothesizing, might provide some benefit on hard problems while harming performance on easy problems.

\vspace{-2mm}

\subsection{Backtracking is a local operator}
\label{sec:backtracking}

We centered the next analysis around \backtracking and \hypothesizing.
Both operators semantically appear to interrupt a model's current line of reasoning, but we asked whether these operators actually reflected global strategic shifts in the model's problem-solving.
We used Claude Sonnet 4.6 as an LLM judge to classify a stratified sample of GPQA events as \textsc{Local} (re-checks a single calculation; method unchanged), \textsc{Sub-Problem} (re-opens a specific case), or \textsc{Global} (abandons the current strategy), prompting the judge with the full reasoning trace and the event marked from the operator span to its next same-operator span.
Local backtracking is explicitly defined as ``The model is re-checking a single calculation, fact lookup, or specific claim...'', and global event is defined as ``model abandons or proposes to replace its current solution...''.
\textbf{\backtracking and \hypothesizing are overwhelmingly local revisions rather than strategy changes}, with statistically indistinguishable scope distributions: \textsc{Backtracking} is 85.4\% \textsc{Local} / 12.8\% \textsc{Sub-Problem} / 1.6\% \textsc{Global} ($n=1{,}294$), and \textsc{Hypothesizing} is 80.2\% / 18.2\% / 1.6\% ($n=192$).
This is supported by literature that shows that backtracking is often superficial \cite{wang2026teaching} and rarely changes a model's answer \cite{kang2025first}. More details are in Appendix~\ref{si:sec:backtracking}.

\vspace{-2mm}
\section{Downstream Applications}
\label{sec:downstream}
\vspace{-2mm}

We now demonstrate how operators discovered by ReasonOps can be used to predict downstream properties of reasoning traces, namely model identity and correctness of the trace.
\vspace{-2mm}
\subsection{Model Identification}
\label{sec:model_id}
\vspace{-2mm}

\begin{figure}[t]
\centering
\includegraphics[width=0.48\textwidth]{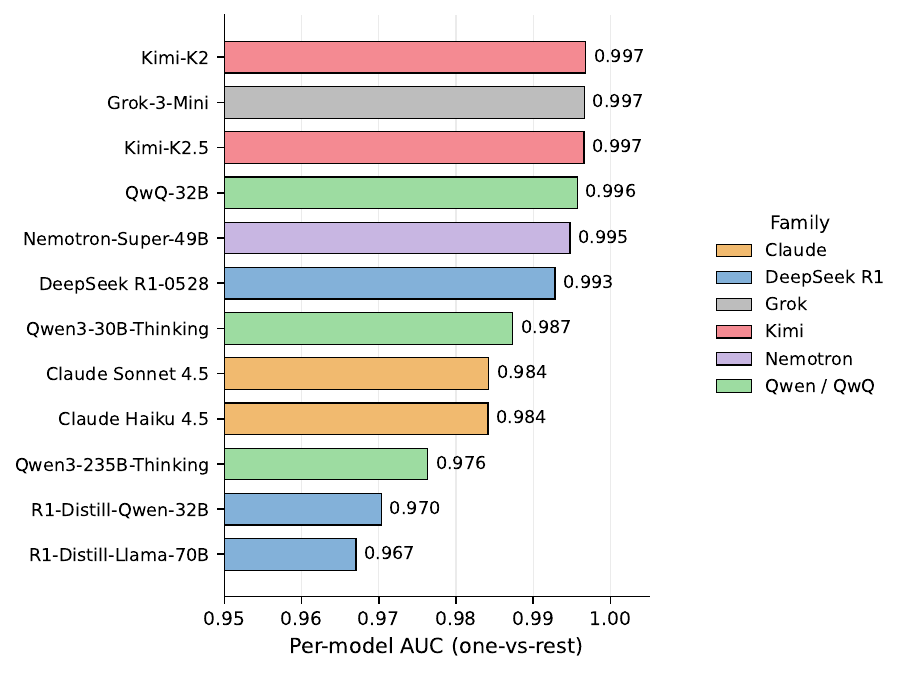}
\caption{Per-model one-vs-rest AUC for the Op-XGB model identification classifier, colored by family. AUCs span $0.967$ (R1-Distill-Llama-70B) to $0.997$ (Kimi-K2, Kimi-K2.5, and Grok-3-Mini, tied); within-family pairs (R1-Distill, Claude, Kimi) sit at the lower end where intra-family confusion is concentrated.}
\label{fig:model_id_barplot}
\end{figure}

If operators capture stable behavioral structure, traces from the same model family should cluster together in operator space. We train \textbf{Op-XGB}, an XGBoost classifier \citep{chen2016xgboost} (400 trees, max depth 6) on a curated operator feature set: a 117-dim handcrafted operator feature vector---global and quartile-localized operator frequencies, bigram transition matrix, run-length statistics, first/last operator one-hots, and entropy/length scalars---concatenated with an 8{,}000-feature anchor-phrase TF-IDF representation; the TF-IDF is re-fit inside each training fold (full feature definitions in Appendix~\ref{app:opxgb}). Op-XGB is trained under problem-level 5-fold CV with model identity as the target. It achieves overall accuracy $= 79.9\%$ (chance $= 8.3\%$) and macro-AUC $= \mathbf{0.987}$. Per-model one-vs-rest AUCs (Figure~\ref{fig:model_id_barplot}) span $0.967$ for R1-Distill-Llama-70B to $0.997$ for Kimi-K2, Kimi-K2.5, and Grok-3-Mini, with the lowest AUCs concentrated in within-family pairs where intra-family confusion is hardest. The full 12-class confusion matrix (Figure~\ref{fig:model_id_confusion}, Appendix~\ref{app:model_id_confusion}) makes this explicit: R1-Distill-Llama-70B traces are predicted as R1-Distill-Qwen-32B at $0.41$ and the reverse at $0.29$ (the largest off-diagonal mass), consistent with their shared distillation lineage from DeepSeek-R1; the Claude-Haiku-4.5 / Claude-Sonnet-4.5 pair shows mutual confusion in the $0.21$--$0.26$ range; and Kimi-K2.5 is predicted as Kimi-K2 at $0.10$ (with the reverse at $0.05$).

\FloatBarrier
\vspace{-2mm}

\subsection{Correctness Prediction}
\label{sec:correctness}
\vspace{-2mm}

We next evaluate whether operator composition can be used to predict answer correctness.
We focus on the 6 benchmarks where operator structure is most informative: AIME, ARC-Challenge, GPQA, LiveCodeBench, MATH-500, and MMLU-Pro. We exclude BIG-Bench Hard and HumanEval: BBH uses answer-matching formats where operator structure has near-chance predictive signal, and HumanEval correctness is determined by code execution rather than reasoning quality. Appendix~\ref{app:per_dataset_si} shows that this exclusion is empirically motivated.
Our primary metric, \textit{WP-AUC} (within-problem AUC; adapted from~\cite{zhou2018din}), computes AUC separately for each problem's trace group and averages---isolating the ability to discriminate better from worse attempts on the \emph{same} problem, with difficulty factored out.
We use 5-fold problem-level CV, assigning all traces for a given problem to the same fold to prevent leakage from multi-model evaluation.\looseness=-1

We compare six methods: (1) \textbf{trace length} (log response character count), (2) \textbf{backtrack count} (fraction of spans labeled \backtracking), (3) \textbf{wait count} (raw count of ``wait''-prefixed spans), (4) \textbf{SelfCheck} \cite{miao2024selfcheck}: the same model that generated the trace reads the problem and its full reasoning and predicts correctness without the gold answer---no training, (5) \textbf{Op-XGB}, and (6) \textbf{OST} (Operator Sequence Transformer; described in \S\ref{sec:early_prediction}). Methods (1)--(3) use logistic regression on a single scalar feature with the same 5-fold problem-level CV protocol; since a 1-D logistic regression is a monotone transformation, AUC is fold-invariant (CD$=$ID). Method (4) requires no training. Methods (5)--(6) are evaluated under two protocols: \textbf{CD} (cross-dataset), trained on all datasets pooled, and \textbf{ID} (in-dataset), trained and evaluated within each dataset separately.

Table~\ref{tab:pred} reports WP-AUC per dataset for all methods.

\begin{table*}[t]
\centering
\caption{Within-problem AUC (WP-AUC) for correctness prediction.
  \textbf{CD} = 5-fold problem-level CV trained on all datasets pooled;
  \textbf{ID} = 5-fold CV trained and evaluated within each dataset.
  Length, Backtrack, and Wait use logistic regression on a single monotonic feature; AUC is rank-invariant to the training fold (CD$=$ID).
    Op-XGB: 117-dim operator features (frequencies, quartile localization, bigram transitions, run lengths) + 8K anchor-phrase TF-IDF, concatenated; XGBoost (400 trees, depth 6). See Appendix~\ref{app:opxgb} for full feature details.
  OST at 100\% trace depth; see \S\ref{sec:early_prediction} for partial-trace results.
  \textbf{Bold} = best. \underline{Underline} = 2nd best. \setlength{\fboxsep}{2pt}\setlength{\fboxrule}{0.5pt}\fcolorbox{lightpurple}{lightpurple}{Our methods.}}
\label{tab:pred}
\small
\setlength{\tabcolsep}{2pt}
\begin{tabular}{l >{\columncolor{lightpurple}}r >{\columncolor{lightpurple}}r >{\columncolor{lightpurple}}r >{\columncolor{lightpurple}}r rr rr rr rr}
    \toprule
    & \multicolumn{2}{>{\columncolor{lightpurple}}c}{Op-XGB} & \multicolumn{2}{>{\columncolor{lightpurple}}c}{OST} & \multicolumn{2}{c}{Length} & \multicolumn{2}{c}{Backtrack} & \multicolumn{2}{c}{Wait} & \multicolumn{2}{c}{SelfCheck} \\
    \cmidrule(lr){2-3}\cmidrule(lr){4-5}\cmidrule(lr){6-7}\cmidrule(lr){8-9}\cmidrule(lr){10-11}\cmidrule(lr){12-13}
    Dataset & CD & ID & CD & ID & CD & ID & CD & ID & CD & ID & CD & ID \\
    \midrule
    AIME          & 0.779 & \textbf{0.838} & \underline{0.801} & 0.797          & 0.500 & 0.500 & 0.621 & 0.621 & 0.679 & 0.679 & 0.501 & 0.501 \\
    ARC-Challenge & \underline{0.618} & 0.559          & \textbf{0.645} & 0.556 & 0.559 & 0.559 & 0.520 & 0.520 & 0.501 & 0.499 & 0.578 & 0.578 \\
    GPQA          & 0.682 & \textbf{0.703} & \underline{0.691} & \underline{0.691} & 0.591 & 0.591 & 0.591 & 0.591 & 0.602 & 0.602 & 0.526 & 0.526 \\
    LiveCodeBench & \underline{0.754} & \textbf{0.795} & 0.727 & 0.732          & 0.426 & 0.574 & 0.616 & 0.616 & 0.591 & 0.591 & 0.496 & 0.496 \\
    MATH          & 0.597 & 0.570          & \underline{0.612} & \textbf{0.662} & 0.407 & 0.407 & 0.464 & 0.464 & 0.483 & 0.517 & 0.504 & 0.504 \\
    MMLU-Pro      & 0.635 & \textbf{0.639} & 0.635 & \underline{0.638}          & 0.587 & 0.587 & 0.628 & 0.628 & 0.633 & 0.633 & 0.520 & 0.520 \\
    \midrule
    \textbf{Global} & 0.701 & \textbf{0.723} & 0.701 & \underline{0.703}        & 0.507 & 0.551 & 0.594 & 0.594 & 0.600 & 0.603 & 0.512 & 0.512 \\
    \bottomrule
\end{tabular}
\end{table*}

Trace length is near chance globally (WP-AUC $= 0.551$ within-dataset). Backtrack count ($0.594$) and wait count ($0.603$) outperform length, confirming that reflective operator frequency carries signal even without full sequence modeling. SelfCheck \cite{miao2024selfcheck}---the same model that generated the trace reading its own reasoning and predicting correctness---reaches WP-AUC $= 0.512$ globally (near chance), a striking null result: reasoning models are overconfident and predict ``correct'' for both correct and incorrect traces at similar rates. Op-XGB reaches $0.701$ cross-dataset and $0.723$ within-dataset globally, with strong within-dataset performance on AIME ($0.838$ ID) and LiveCodeBench ($0.795$ ID). The OST reaches $0.701$ cross-dataset and $0.703$ within-dataset at full trace depth---tied with Op-XGB cross-dataset despite reading only operator labels, with no access to anchor phrase text. OST surpasses Op-XGB on ARC-Challenge ($0.645$ vs.\ $0.618$ CD) and MATH ($0.662$ vs.\ $0.570$ ID), where temporal operator dynamics capture structure that aggregate feature counts miss.

\vspace{-2mm}

\FloatBarrier
\subsection{Early Correctness Prediction}
\label{sec:early_prediction}
\vspace{-2mm}

\begin{figure}[t]
\centering
\includegraphics[width=\linewidth]{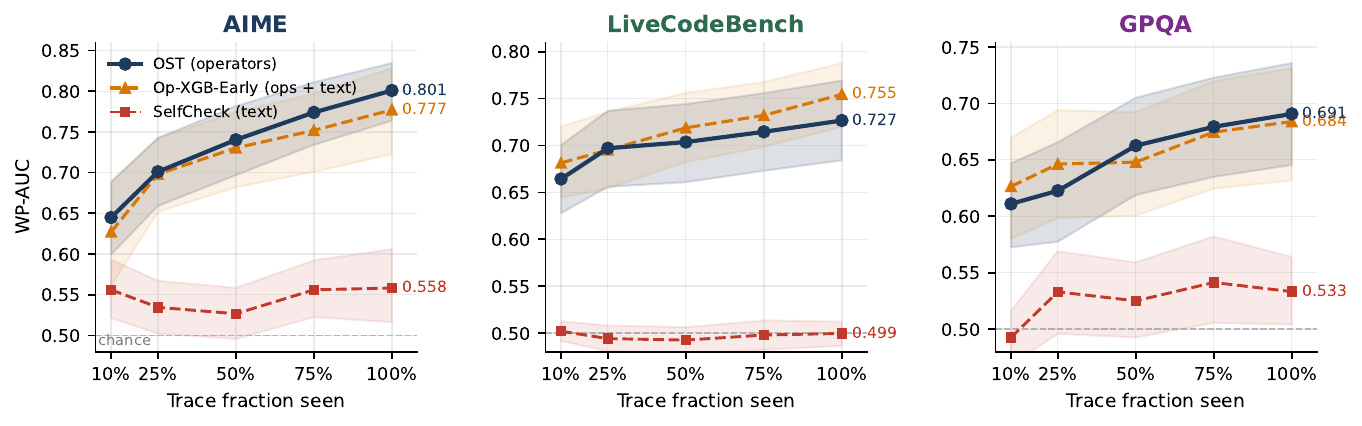}
\caption{Correctness prediction WP-AUC at increasing trace depths for the three best-performing datasets (AIME, LiveCodeBench, GPQA). Navy solid (\textbf{OST}): operator sequence transformer trained once on full sequences, reading the discrete operator label sequence at any prefix length without retraining. Amber dashed (\textbf{Op-XGB-Early}): full Op-XGB recipe (117-dim operator features + 8{,}000-dim anchor TF-IDF; Appendix~\ref{app:opxgb}), retrained from scratch at each depth as an upper-bound comparison. Red dashed (\textbf{SelfCheck}): the same model that generated the trace reads its own raw reasoning up to depth $p\%$ and predicts correctness with no training. Shaded bands: 95\% CIs.}
\label{fig:early_prediction}
\end{figure}

A practical question is whether correctness can be estimated from a partial trace, before the model finishes reasoning. We introduce the \textbf{Operator Sequence Transformer} (OST): a lightweight Transformer encoder (${\sim}800$K parameters; 4 layers, $d{=}128$, 4 heads, pre-LayerNorm) trained end-to-end directly on operator sequences with no pretraining. Each operator token is represented as the sum of a unigram embedding, a \emph{bigram} (transition) embedding encoding the previous$\to$current operator pair, and a continuous sinusoidal position encoding proportional to the token's normalized position within the visible prefix---giving the model a fine-grained sense of \emph{where} in the trace each operator appears. The model is trained with a pure contrastive loss that directly encourages correct traces to score higher than incorrect traces on the same problem, under the same 5-fold problem-level CV protocol as the other methods. Crucially, the OST is \emph{trained once on full operator sequences}; at inference time it accepts any operator-sequence prefix without retraining, making it a practical single-model early-prediction system deployable at any trace depth. More details are in Appendix~\ref{app:ost}.

Figure~\ref{fig:early_prediction} shows OST WP-AUC vs.\ trace depth for the three best-performing datasets.
On AIME, WP-AUC rises from $0.645$ at 10\% depth to $0.740$ at 50\% and $0.801$ at 100\%---at half the trace it already surpasses all span-free baselines.
LiveCodeBench and GPQA follow similar trajectories, reaching $0.727$ and $0.691$ at full depth.
Globally, the OST rises from WP-AUC $= 0.605$ at 10\% depth to $0.664$ at 50\% and $0.701$ at 100\%, monotonically improving with trace length.

We compare against \textbf{Op-XGB-Early}, which applies the same Op-XGB feature recipe at each depth $d$: we truncate every trace to its first $d\%$ of spans, recompute the operator-sequence statistics and the TF-IDF on the truncated span set, and fit a fresh XGBoost from scratch. Per-depth retraining gives Op-XGB-Early an explicit upper-bound advantage---it is re-optimized for each truncation level---whereas the OST is trained only once on full sequences. Globally, Op-XGB-Early reaches WP-AUC $0.620$ at 10\%, $0.647$ at 25\%, $0.661$ at 50\%, $0.682$ at 75\%, and $0.699$ at 100\%. The OST and Op-XGB-Early track each other within a few thousandths across all depths: OST$-$Op-XGB-Early gaps are $-0.015$, $-0.016$, $+0.003$, $+0.006$, and $+0.003$ at depths $\{10, 25, 50, 75, 100\}\%$. The single OST model thus matches a per-depth-retrained Op-XGB---which has access to both the operator features \emph{and} the raw anchor text---for partial-trace prediction at 50\% depth and beyond, despite reading only the discrete operator label sequence and never being retrained for partial inputs.

By contrast, the partial-trace SelfCheck baseline---the same model reading its own first $p\%$ of raw reasoning text---is near chance at all depths ($0.50$--$0.56$ across datasets), confirming that the OST's signal comes from the structural operator representation rather than from text content alone.

\vspace{-2mm}

\FloatBarrier
\section{Discussion}
\label{sec:discussion}
\vspace{-2mm}

ReasonOps is an unsupervised method for annotating reasoning traces, built from careful observations of their semantic structure.
Three-token pivots strike a useful balance for discourse analysis: they disambiguate the polysemy of single-token markers like "but" or "so" (e.g., distinguishing "wait," "wait no," and "wait let me" as backtracking, rejection, and correction respectively) while remaining short enough to recur across tens of thousands of traces. The top-2,000 vocabulary filter is essential to this approach, as it excludes high-frequency but domain-specific phrases (e.g., "the double bond" in chemistry traces) that would otherwise contaminate the pivot set with non-discourse-functional content.
ReasonOps is a robust method to provide domain-agnostic, model-agnostic annotations of reasoning traces that enable universal analysis and downstream tasks such as model identification and correctness prediction.

\xhdr{Future work and limitations} We see several avenues for future work on annotating reasoning traces.
An extension could be made to agentic traces, i.e., traces of actions taken by an LLM when enabled with a harness.
The ReasonOps approach, which is unsupervised and does not assume \textit{a priori} sets of operators, might be more useful for agentic traces where actions may change drastically based on the harness.
The operators define a structured vocabulary for step-level correctness prediction without human annotation. OST's early correctness signal can trigger intervention (e.g., beam expansion, early stopping) at any trace position, enabling compute-adaptive generation. There are also several limitations of ReasonOps. First, the pipeline relies on sentence-initial pivots; spans bounded at a sub-sentence level cannot be labeled. Second, the operators are validated by LLM judgment, which introduces its own biases. Third, cross-model transfer is limited: operator frequency features partially capture model-specific patterns that do not fully generalize. Finally, the corpus covers 8 benchmarks in English; generalization to other languages is not verified.

\xhdr{Broader impacts} This work paves the way to a common language for describing reasoning traces of LLMs. This could be used for further analysis of LLMs as well as downstream tasks beyond correctness prediction.

\vspace{-2mm}

\paragraph{Summary.}
Seven reasoning operators emerge from unsupervised analysis of 44{,}662 chain-of-thought traces across 12 thinking LLMs and 8 benchmarks. Three independent LLM judges confirm the taxonomy (70--76\% classification accuracy; chance: 14\%). Structural operator features plus anchor-phrase features identify the source model with macro-AUC $= 0.987$. Structural features predict within-problem correctness well above all span-free baselines and above the LLM self-judge ($0.512$, near chance). The OST reaches WP-AUC $= 0.701$ cross-dataset---tied with Op-XGB ($0.701$ CD) while reading only operator labels, not the additional text features Op-XGB receives. The OST predicts correctness monotonically from partial traces, reaching $0.664$ at 50\% trace depth and surpassing an Op-XGB baseline retrained at each depth, with no per-depth retraining required. The convergence of architecturally diverse models on the same 7 discourse-level moves suggests that extended reasoning organizes itself into a common compositional vocabulary.

\section*{Acknowledgments}
This work was supported in part by computational resources and research infrastructure from the Stanford AI Lab.
We thank colleagues at Stanford, including Grant Wilkins, Elana Simon, and members of the Zou lab for helpful discussions and feedback.

\bibliography{references}

\begin{thebibliography}{8}

\bibitem{qwq32b}
{Qwen Team}.
\newblock {QwQ-32B}: embracing the power of reinforcement learning.
\newblock \url{https://qwenlm.github.io/blog/qwq-32b/}, 2025.
\newblock Blog post, March 6, 2025.

\bibitem{yang2025qwen3S}
An Yang, Anfeng Li, Baosong Yang, Beichen Zhang, Binyuan Hui, Bo Zheng, Bowen
  Yu, Chang Gao, Chengen Huang, Chenxu Lv, et~al.
\newblock {Qwen3} technical report.
\newblock {\em arXiv preprint arXiv:2505.09388}, 2025.

\bibitem{deepseek2025r1}
{DeepSeek-AI}, Daya Guo, Dejian Yang, Haowei Zhang, Junxiao Song, Peiyi Wang,
  Qihao Zhu, et~al.
\newblock {DeepSeek-R1}: incentivizing reasoning capability in {LLMs} via
  reinforcement learning.
\newblock {\em arXiv preprint arXiv:2501.12948}, 2025.

\bibitem{anthropic2025claude}
{Anthropic}.
\newblock Introducing {Claude} 4.
\newblock \url{https://www.anthropic.com/news/claude-4}, 2025.
\newblock Blog post, May 22, 2025.

\bibitem{xai2025grok3}
{xAI}.
\newblock Grok 3 beta --- the age of reasoning agents.
\newblock \url{https://x.ai/news/grok-3}, 2025.
\newblock Blog post, February 19, 2025.

\bibitem{nvidia2025nemotron}
Akhiad Bercovich, Itay Levy, Izik Golan, Mohammad Dabbah, Ran El-Yaniv, Omri
  Puny, et~al.
\newblock Llama-nemotron: efficient reasoning models.
\newblock {\em arXiv preprint arXiv:2505.00949}, 2025.

\bibitem{kimiteam2025k2}
{Kimi Team}.
\newblock Kimi k2: open agentic intelligence.
\newblock {\em arXiv preprint arXiv:2507.20534}, 2025.

\bibitem{team2026kimiS}
Kimi Team, Tongtong Bai, Yifan Bai, Yiping Bao, S.~H. Cai, Yuan Cao, Y.~Charles,
  H.~S. Che, Cheng Chen, Guanduo Chen, et~al.
\newblock Kimi k2.5: visual agentic intelligence.
\newblock {\em arXiv preprint arXiv:2602.02276}, 2026.

\end{thebibliography}
\bibliographystyle{unsrt}

\clearpage
\appendix

\section{Model Details}
\label{app:models}

Table~\ref{tab:models} lists the 12 thinking models used for trace collection. All expose raw chain-of-thought tokens accessible via API. All models except Claude are queried via the OpenRouter API; reasoning tokens are returned in the \texttt{reasoning} field of the OpenRouter response message. The \texttt{reasoning} request parameter is set to \texttt{\{effort: high\}} for Grok-3-Mini, \texttt{\{max\_tokens: N\}} for Qwen3 thinking variants, and omitted for DeepSeek, QwQ-32B, Kimi, and Nemotron models. Claude models are accessed directly via the Anthropic API with \texttt{thinking: \{type: enabled, budget\_tokens: N\}} enabled; raw thinking content is extracted from \texttt{type: thinking} response blocks.

\begin{table}[ht]
\centering
\caption{Models used for trace collection. All 12 expose raw chain-of-thought tokens.}
\label{tab:models}
\small
\begin{tabular}{llrl}
\toprule
Model & Family & Size & Notes \\
\midrule
QwQ-32B              & Qwen    & 32B            & S\citeS{qwq32b} \\
Qwen3-30B-Thinking   & Qwen    & 30B (3B act.)  & MoE S\citeS{yang2025qwen3S} \\
Qwen3-235B-Thinking  & Qwen    & 235B (22B act.)& MoE S\citeS{yang2025qwen3S} \\
DeepSeek-R1-0528     & DeepSeek& 671B MoE       & S\citeS{deepseek2025r1} \\
R1-Distill-LLaMA-70B & DeepSeek& 70B            & Distilled S\citeS{deepseek2025r1} \\
R1-Distill-Qwen-32B  & DeepSeek& 32B            & Distilled S\citeS{deepseek2025r1} \\
Claude-Sonnet-4.5    & Anthropic& ---            & Extended thinking S\citeS{anthropic2025claude} \\
Claude-Haiku-4.5     & Anthropic& ---            & Extended thinking S\citeS{anthropic2025claude} \\
Grok-3-Mini          & xAI     & ---            & S\citeS{xai2025grok3} \\
Nemotron-Super-49B   & NVIDIA  & 49B            & S\citeS{nvidia2025nemotron} \\
Kimi-K2-Thinking     & Moonshot& MoE            & S\citeS{kimiteam2025k2} \\
Kimi-K2.5            & Moonshot& MoE            & S\citeS{team2026kimiS} \\
\bottomrule
\end{tabular}
\end{table}

\section{Corpus Statistics}
\label{app:corpus}

The full corpus spans 8 benchmarks and contains 44,662 traces across 12 models. Table~\ref{tab:corpus} reports per-dataset statistics for the 6-dataset correctness prediction subset (33,209 traces), which excludes BBH and HumanEval due to grading ambiguity or near-ceiling accuracy. Each problem was attempted by all 12 models with up to 5 independent samples per model; the AIME trace count is lower because AIME 2024 has only 30 problems.

\begin{table}[ht]
\centering
\caption{Per-dataset trace counts in the 6-dataset correctness prediction corpus. All 12 models contribute to every dataset. Accuracy = fraction of traces judged correct by the official grader.}
\label{tab:corpus}
\small
\begin{tabular}{lrrrl}
\toprule
Dataset & Traces & Problems & Models & Accuracy \\
\midrule
AIME 2024         & 4{,}604 &  90 & 12 & 86.4\% \\
ARC-Challenge     & 5{,}808 &  97 & 12 & 94.9\% \\
GPQA Diamond      & 5{,}615 & 100 & 12 & 74.0\% \\
LiveCodeBench     & 5{,}516 &  99 & 12 & 33.5\% \\
MATH-500          & 5{,}794 & 100 & 12 & 94.6\% \\
MMLU-Pro          & 5{,}872 & 100 & 12 & 76.3\% \\
\midrule
\textbf{Total}    & \textbf{33{,}209} & \textbf{586} & & \\
\bottomrule
\end{tabular}
\end{table}

The full operator-discovery corpus additionally includes BIG-Bench Hard and HumanEval (44{,}662 traces total across all 8 benchmarks). These two datasets are excluded from correctness prediction: BBH uses answer-matching formats where operator structure shows near-chance predictive signal, and HumanEval correctness is determined by code execution rather than reasoning quality.

\section{\texorpdfstring{$K$}{K} Selection}
\label{app:ksweep}

Table~\ref{tab:ksweep} reports LLM-judge $\kappa$ (Claude Sonnet~4.6) for each $K$ in the sweep. $K{=}7$ achieves the highest $\kappa$.

\begin{table}[ht]
\centering
\caption{$K$ selection via LLM-judge $\kappa$ (Claude Sonnet~4.6).}
\label{tab:ksweep}
\small
\begin{tabular}{rr}
\toprule
$K$ & $\kappa$ \\
\midrule
6 & 0.604 \\
\textbf{7} & \textbf{0.693} \\
8 & 0.611 \\
9 & 0.648 \\
10 & 0.573 \\
11 & 0.558 \\
\bottomrule
\end{tabular}
\end{table}

\section{Operator Sequence Transformer (OST) Architecture}
\label{app:ost}

The OST is a lightweight Transformer encoder trained end-to-end on operator token sequences. Its design is optimized for long sequences of discrete operator labels (typical length: 50--500 tokens) while remaining parameter-efficient enough to train in 5-fold CV.

\paragraph{Tokenization.}
Each span in a trace is assigned one of 7 operator labels (integer indices $0$--$6$). The OST processes the sequence of these labels for a trace prefix of specified depth.

\paragraph{Input embeddings.}
Each position $t$ in the operator sequence receives a 128-dimensional input vector formed as the sum of three components:
\begin{itemize}
  \item \textbf{Unigram embedding}: a learned embedding $E_u[o_t] \in \mathbb{R}^{128}$ for the current operator $o_t$;
  \item \textbf{Bigram (transition) embedding}: a learned embedding $E_b[o_{t-1}, o_t] \in \mathbb{R}^{128}$ for the previous$\to$current operator pair (special ``start'' token for $t=0$);
  \item \textbf{Continuous sinusoidal position encoding}: standard sinusoidal encoding scaled by the \emph{normalized} position $t / T$ within the visible prefix, giving the model a fine-grained sense of where in the trace each operator appears.
\end{itemize}

\paragraph{Transformer body.}
The model uses a standard pre-LayerNorm Transformer encoder with 4 layers, $d_\text{model} = 128$, 4 attention heads, feed-forward width $= 512$, and $5\%$ token dropout during training. Attention is full (not causal), since the model reads a complete prefix.

\paragraph{Pooling and scoring.}
Sequence representations are aggregated by attention pooling (learned query vector) into a single 128-dimensional vector, followed by a two-layer MLP ($128 \to 64 \to 1$) producing a real-valued correctness score. Total parameter count: $\approx$800{,}000.

\paragraph{Training.}
The OST is trained with a pure contrastive loss: for every problem with at least one correct and one incorrect trace in the training fold, all pairs $(i^+, i^-)$ of correct and incorrect traces for that problem are formed, and the loss pushes the correct trace to score higher. This directly optimizes the within-problem ranking that WP-AUC measures. Training uses AdamW ($\eta = 3 \times 10^{-4}$, weight decay $= 0.01$), batch size 64, and early stopping on fold validation WP-AUC.

\paragraph{Early prediction.}
At inference time the OST is given only the first $p\%$ of operator tokens (by span count), allowing evaluation at any trace depth. The sinusoidal position encoding is re-normalized to the partial prefix length.

\section{Op-XGB Feature Details}
\label{app:opxgb}

Op-XGB combines a 117-dimensional handcrafted operator-sequence feature vector with an 8,000-feature anchor-phrase TF-IDF representation. Both feature sets are derived from the operator labels and sentence-initial pivot phrases---no substantive math, code, or reasoning content of the trace is used.

\paragraph{Operator feature vector (117 dim).}
For each trace's operator sequence $\sigma = (\sigma_1, \dots, \sigma_n) \in \{0,\dots,6\}^n$:
\begin{itemize}\itemsep0pt
  \item \textbf{Global frequencies} (7): the share of each operator across the full sequence, $\mathrm{freq}_k = \tfrac{1}{n}\sum_t \mathbf{1}[\sigma_t = k]$.
  \item \textbf{Quartile frequencies} ($7 \times 4 = 28$): the share of each operator within each of four equal-sized contiguous chunks of $\sigma$, capturing temporal localization.
  \item \textbf{Scalar summaries} (5): Shannon entropy of the global frequency vector, the maximum of $\mathrm{freq}$, the number of distinct operators that appear, $\log(1 + n)$, and the immediate-repeat rate $\tfrac{1}{n-1}\sum_{t \ge 2}\mathbf{1}[\sigma_t = \sigma_{t-1}]$.
  \item \textbf{First/last one-hot} ($7 + 7 = 14$): one-hot encodings of $\sigma_1$ and $\sigma_n$.
  \item {\sloppy\textbf{Bigram transition matrix} (49): the $7 \times 7$ matrix of bigram frequencies $\tfrac{1}{n-1}\sum_{t \ge 2}\mathbf{1}[\sigma_{t-1}{=}a, \sigma_t{=}b]$, flattened to a 49-dimensional vector of normalized transition counts.\par}
  \item \textbf{Run-length statistics} ($7 + 7 = 14$): for each operator $k$, the mean and max length of consecutive runs of $k$, each normalized by $n$.
\end{itemize}
This vector is computed once per trace and depends only on the discrete operator label sequence; it carries no anchor text.

\paragraph{Anchor-phrase TF-IDF (8{,}000 dim).}
We additionally extract each span's anchor (the sentence-initial pivot phrase, truncated to 80 characters), concatenate all anchors of a trace with single spaces, and fit \texttt{sklearn.feature\_extraction.text.TfidfVectorizer} with \texttt{max\_features=8000} and \texttt{sublinear\_tf=True} on the training-fold anchor strings (default unigram tokenization, sublinear TF $1 + \log(\mathrm{tf})$, IDF reweighting, vocabulary re-fit inside each fold to prevent leakage). The anchor strings consist only of discourse-level pivot phrases (e.g., ``let's denote'', ``so we have'', ``wait, that means'')---they are exactly the inputs that the unsupervised operator-discovery pipeline embeds and clusters into the 7 operators.

\paragraph{Classifier.}
The 117-dim operator feature vector and the 8{,}000-dim TF-IDF vector are concatenated to form an 8{,}117-dim per-trace representation, which is passed to XGBoost (400 trees, max depth 6, learning rate 0.05, subsample 0.8, colsample\_bytree 0.8, logloss objective). The classifier is trained with problem-level 5-fold CV (cross-dataset) or within each dataset separately (in-dataset). XGBoost predicted probabilities are used directly as scores for WP-AUC computation. Op-XGB therefore reads operator information at \emph{both} levels of abstraction (discrete labels and the raw pivot phrases that produce them), in contrast to the OST, which reads only the discrete labels.

\paragraph{Op-XGB-Early variant.}
For the early-prediction comparison (\S\ref{sec:early_prediction}, Figure~\ref{fig:early_prediction}), at each depth $d \in \{10, 25, 50, 75, 100\}\%$ we truncate each trace's spans to the first $\lceil N \cdot d / 100 \rceil$ spans, recompute the 117-dim operator feature vector on the truncated operator subsequence, recompute the anchor TF-IDF on the truncated anchor text, and re-fit a fresh XGBoost on the concatenated representation under the same 5-fold problem-level CV. This per-depth retraining provides an upper bound on what the full Op-XGB feature recipe can achieve when the model is re-optimized for each truncation level---in contrast to the OST, which is trained once on full sequences and evaluated on partial prefixes without retraining.

\paragraph{Op-XGB for model identification.}
For 12-class source-model identification (Figure~\ref{fig:model_id_confusion}), we use the identical Op-XGB feature recipe described above (117-dim operator features + 8{,}000-dim TF-IDF, concatenated to an 8{,}117-dim vector) and train XGBoost as a multi-class classifier (\texttt{multi:softprob}, same hyperparameters as the binary version) under the same problem-level 5-fold CV protocol.
Reusing the full Op-XGB recipe ensures that the same model class supports both correctness prediction and model identification.

\section{Operator Usage by Correctness per Benchmark}
\label{app:correctness_heatmap}

\begin{figure}[ht]
\centering
\includegraphics[width=\linewidth]{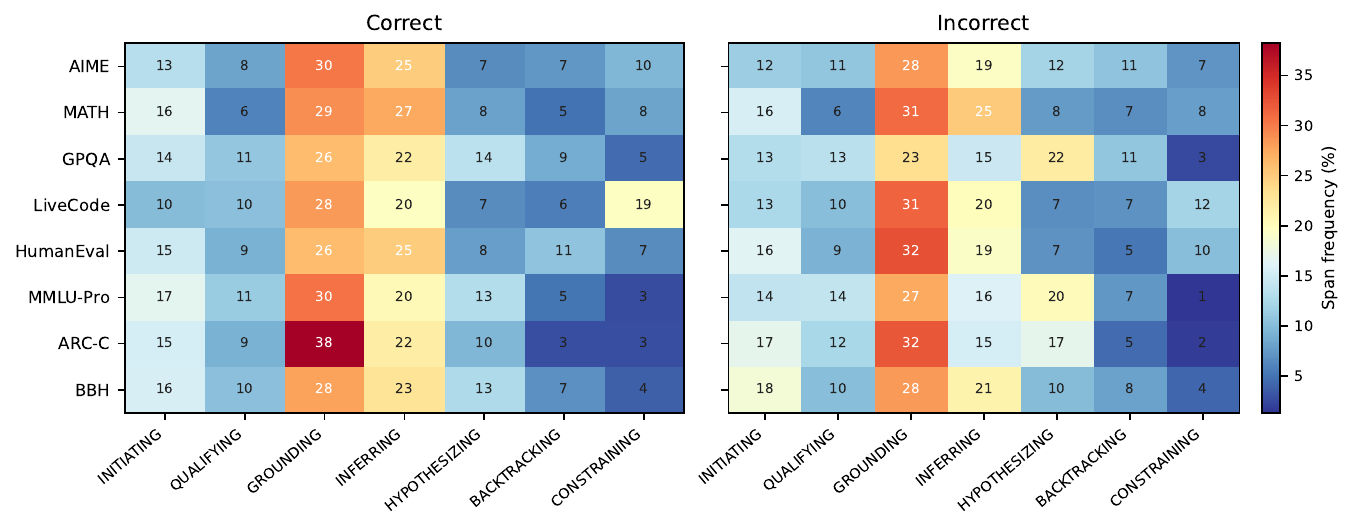}
\caption{Operator frequency (\%) per benchmark, split by correctness. Left: correct traces. Right: incorrect traces.}
\label{fig:dataset_heatmap_correctness}
\end{figure}

Figure~\ref{fig:dataset_heatmap_correctness} shows operator frequency separately for correct and incorrect traces. \textsc{Inferring} is consistently higher in correct traces; \textsc{Hypothesizing} and \textsc{Backtracking} are higher in incorrect traces. The gap is largest on commonsense benchmarks (ARC-C, MMLU-Pro: $\sim$24\% incorrect vs.\ $\sim$14\% correct for \textsc{Hypothesizing}). On LiveCodeBench, \textsc{Constraining} is 10 pp higher in correct traces.

\section{Temporal Operator Profiles}
\label{app:temporal}

\begin{figure}[ht]
\centering
\includegraphics[width=\linewidth]{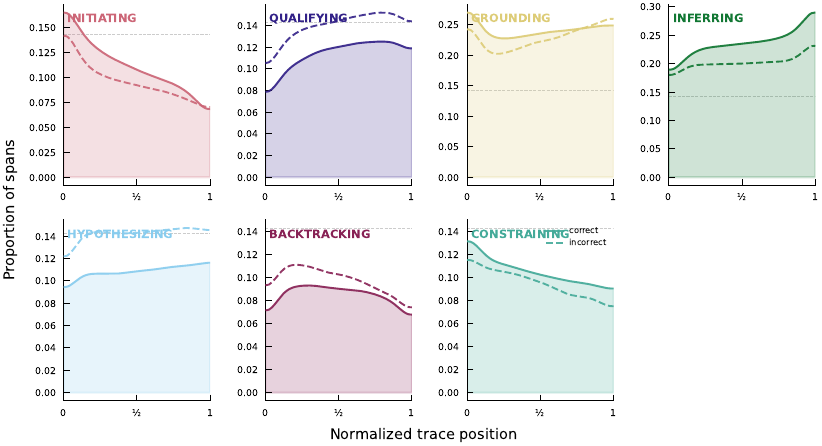}
\caption{Operator density as a function of normalized trace position (Gaussian KDE). Dashed reference = uniform (1/7). Correct: solid; incorrect: dashed.}
\label{fig:temporal}
\end{figure}

Figure~\ref{fig:temporal} shows operator density as a function of normalized trace position. \textsc{Grounding} dominates the early trace; \textsc{Inferring} concentrates near the end. \textsc{Backtracking} shows a midtrace spike in incorrect traces---a signature of failed exploration not followed by recovery.

\clearpage

\section{Backtracking analysis details}
\label{si:sec:backtracking}

We classify each \textsc{Backtracking} and \textsc{Hypothesizing} event on GPQA as \textsc{Local} (re-checks a single calculation; method unchanged), \textsc{Sub-Problem} (re-opens a specific case or branch), or \textsc{Global} (abandons the current solution strategy), using Claude Sonnet 4.6 as a judge.

\paragraph{Sample.}
Events are drawn from GPQA traces with stratification over (model, correctness), yielding $n{=}1{,}294$ \textsc{Backtracking} and $n{=}192$ \textsc{Hypothesizing} events.

\paragraph{Prompt.}
The judge sees the GPQA problem and the full reasoning trace with the target event marked: a contiguous region from the target operator span to the next span of the same operator class (or trace end). The system rubric is reproduced below; the judge returns a JSON object \texttt{\{"scope": ..., "rationale": ...\}}. See prompt below for more information.

\begin{table}[ht]
\centering
\begin{tabular}{lcccc}
\hline
Operator & Local & Sub-Problem & Global & $n$ \\
\hline
\textsc{Backtracking} & 85.4\% & 12.8\% & 1.6\% & 1,294 \\
\textsc{Hypothesizing} & 80.2\% & 18.2\% & 1.6\% & 192 \\
\hline
\end{tabular}
\caption{Scope classification for \textsc{Backtracking} and \textsc{Hypothesizing} events on GPQA.}
\label{tab:scope_results}
\end{table}

\begin{tcolorbox}[
    breakable,
    enhanced,
    title={Prompt: Backtracking analysis judge prompt},
    fonttitle=\bfseries,
    colback=gray!4,
    colframe=black!70,
    boxrule=0.4pt,
    arc=2pt,
    left=6pt, right=6pt, top=6pt, bottom=6pt,
]
\small
You are analyzing reasoning traces from large language models solving GPQA (graduate-level science) questions. Your task is to classify the SCOPE of a single REVISION-ADJACENT region marked in the trace---a contiguous section where the model re-checks, considers an alternative, hypothesizes, qualifies, or hesitates.

The marked region (between \verb!>>>MARKER<<<! and \verb!<<<END>>>! markers) starts at the target operator event and ends just before the next event of the SAME operator type (or at the end of the trace). Read the surrounding context to decide what the model is actually re-thinking inside the marked region.

\medskip
\noindent\textbf{CLASSIFY THE SCOPE:}

\smallskip
\noindent\textbf{LOCAL:}\ The model is re-checking a single calculation, fact lookup, or specific claim (e.g., ``wait, $3 \times 7 = 21$ not $18$''; ``actually that should be oxygen''; ``let me recompute that integral''; ``I think this is the methyl group''). The OVERALL APPROACH is unchanged---same method, same plan, just verifying or fixing one step. Includes self-confirmation moves and tentative single-fact hypotheses.

\smallskip
\noindent\textbf{SUB\_PROBLEM:}\ The model is re-opening or alternating among specific cases, branches, or sub-questions within the larger problem (e.g., ``let me reconsider case 2''; ``what about the limit where $x \to 0$?''; ``could this be the (R) instead of (S) configuration?''). Strategy preserved but a discrete piece is redone or swapped.

\smallskip
\noindent\textbf{GLOBAL:}\ The model abandons or proposes to replace its current solution strategy/method with a fundamentally different approach (e.g., ``this Lagrangian approach isn't working, let me try energy conservation''; ``I shouldn't use the ideal-gas assumption---let me redo with van der Waals''; ``I had the wrong mechanism entirely''; ``let me use Gauss's law instead of Coulomb''). The high-level plan changes.

\medskip
\noindent\textbf{Output format:} a single JSON object on one line:
\begin{center}
\verb!{"scope": "LOCAL"|"SUB_PROBLEM"|"GLOBAL", "rationale": "<one short sentence>"}!
\end{center}

Be decisive. Compare what the model is doing in the spans IMMEDIATELY before vs.\ after the marked region---that local context tells you what is actually changing. Judge each case on its merits without a default fallback category.
\tcblower
\captionsetup{hypcap=false}
\captionof{prompt}{System prompt used to classify the scope of \backtracking and \hypothesizing spans in GPQA reasoning traces in backtracking analysis.}
\end{tcolorbox}

\clearpage

\section{Per-Dataset Baseline Comparison}
\label{app:per_dataset_si}

\begin{figure}[ht]
\centering
\includegraphics[width=\linewidth]{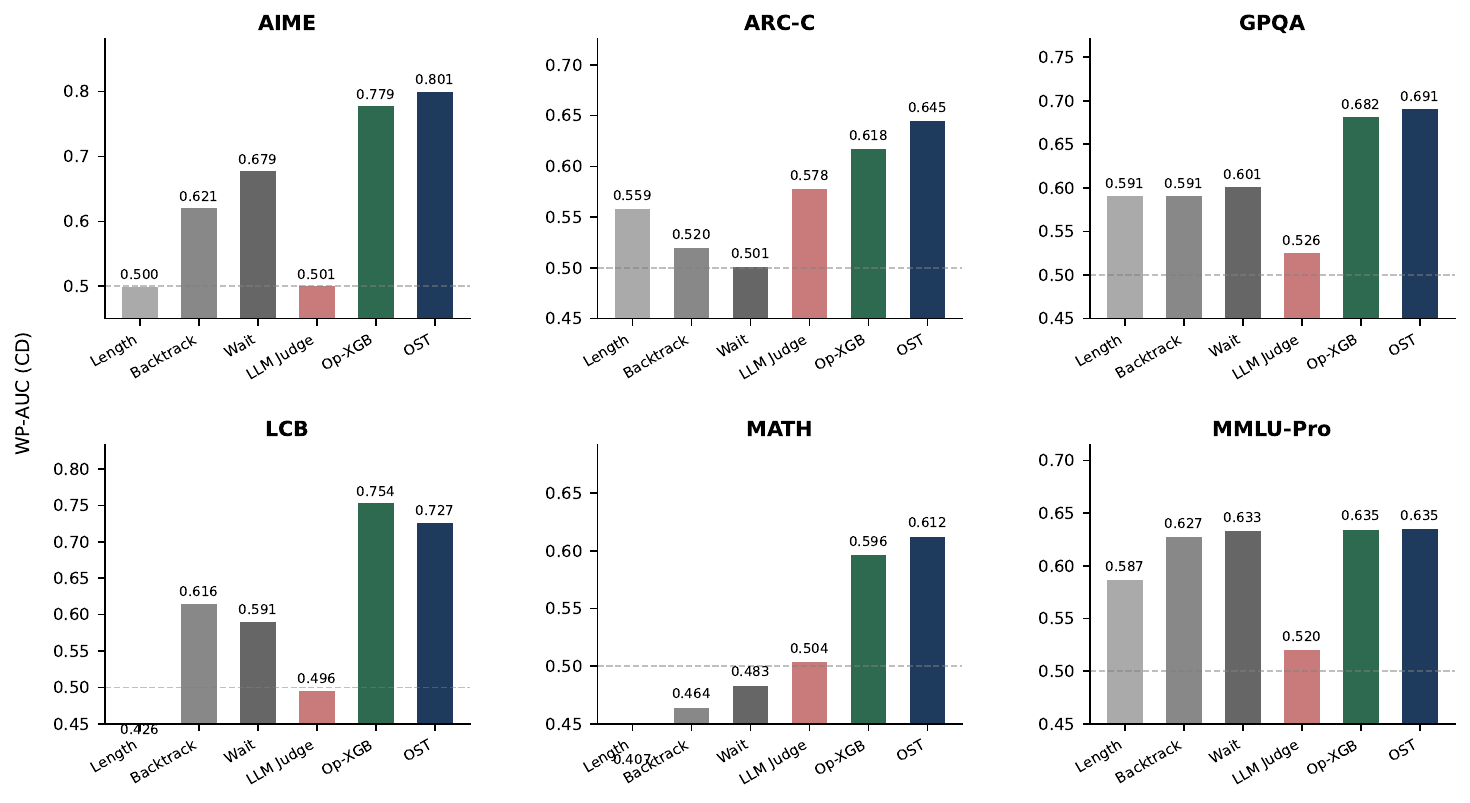}
\caption{Per-benchmark WP-AUC (cross-dataset CV) for all six methods. Gray bars: span-free baselines (Length, Backtrack, Wait). Red bar: LLM self-judge. Green bar: Op-XGB (operator n-gram counts + anchor-phrase TF-IDF; XGBoost). Blue bar: OST (Operator Sequence Transformer).}
\label{fig:per_dataset_wpauc}
\end{figure}

Figure~\ref{fig:per_dataset_wpauc} shows WP-AUC (cross-dataset protocol) across all six baselines for each of the 6 benchmarks. BIG-Bench Hard and HumanEval are excluded (see \S\ref{sec:correctness} and Appendix~\ref{app:corpus}). Op-XGB and OST dominate on math-heavy and code benchmarks; the LLM self-judge is near chance on all datasets.

\section{Labeled Trace Example}
\label{sec:appendix}

Table~\ref{tab:trace_example} shows a representative labeled trace from R1-Distill-LLaMA-70B on a MATH intermediate algebra problem (incorrect). Each span begins at a pivot sentence (anchor, upright font) followed by continuation sentences in \textit{italics}.

\begin{table}[ht]
\centering
\caption{Labeled trace with full span content. Model: R1-Distill-LLaMA-70B. Dataset: MATH. Correct: False.}
\label{tab:trace_example}
\small
\begin{tabular}{lp{9cm}}
\toprule
\multicolumn{2}{p{10.5cm}}{\textit{Problem:} Find all integer roots of $x^4{+}5x^3{+}9x^2{-}x{-}14{=}0$.} \\[4pt]
\midrule
Operator & Span content \\
\midrule
\textsc{Inferring}     & Okay, so I need to solve this equation and find all the integer roots. \newline \textit{And the task is to find all the integer roots.} \\[3pt]
\textsc{Hypothesizing} & Hmm, I remember that for polynomials the Rational Root Theorem can be helpful. \\[3pt]
\textsc{Initiating}    & Let me try to recall how that works. \newline \textit{The Rational Root Theorem states that any rational root $p/q$ has $p\mid a_0$ and $q\mid a_n$.} \\[3pt]
\textsc{Hypothesizing} & Therefore, the possible integer roots are the factors of $-14$. \\[3pt]
\textsc{Initiating}    & Let me start with $x{=}1$ and $x{=}{-1}$ first. \\[3pt]
\textsc{Grounding}     & Substituting: $1{+}5{+}9{-}1{-}14{=}0$. \newline \textit{So $x{=}1$ is a root.} \\[3pt]
\textsc{Inferring}     & So $x{=}{-2}$ is another integer root. \\[3pt]
\textsc{Backtracking}  & Wait, but let me double-check my calculations because sometimes I might make an arithmetic mistake. \\[3pt]
\textsc{Grounding}     & Starting with $x{=}1$: $1{+}5{+}9{-}1{-}14{=}0$. Correct. \\[3pt]
\textsc{Backtracking}  & But wait, the equation is a quartic, so it should have four roots in total. \\[3pt]
\textsc{Qualifying}    & But according to our tests, the other candidates didn't work, so perhaps the other roots are non-integer. \\[3pt]
\textsc{Hypothesizing} & Alternatively, maybe I missed something, so let me try another approach. \\[3pt]
\textsc{Constraining}  & Now, we have factored the polynomial as $(x{-}1)(x^3{+}6x^2{+}15x{+}14)$. \\[3pt]
\textsc{Qualifying}    & Therefore, the only integer roots are $1$ and $-2$. \\
\bottomrule
\end{tabular}
\end{table}

\section{Evaluation}

\subsection{Stability Analysis}
\label{app:stability}

\paragraph{Methodology.}
To assess whether the induced operator labels reflect stable semantic
structure rather than arbitrary exemplar choice, we held the $K=7$
cluster assignments fixed and re-ran the LLM naming step
(\texttt{claude-sonnet-4-6}, $T=0$; same prompt as the discovery
pipeline---top-12 anchor $n$-grams plus 5 example spans per
cluster) across 30 exemplar seeds. The top $n$-grams are deterministic
per cluster; only the 5 exemplars vary by seed, isolating naming
stability from the upstream clustering. Each of the
$30 \times 7 = 210$ resulting descriptions was embedded once with
\texttt{text-embedding-3-small} (1536-d, $L_2$-normalized), and we
computed pairwise cosine similarity as the inner product. We then
formed three pair distributions: a \emph{within-real} distribution
(all $\binom{30}{2} = 435$ cross-seed pairs per cluster, $3{,}045$ pairs
pooled across the seven operators); an \emph{across-real baseline} of
$2{,}000$ random pairs $(i, j)$ with $\mathrm{cluster}(i) \neq
\mathrm{cluster}(j)$ from the same description pool; and a
\emph{random-group control} generated by the identical naming protocol
applied to seven disjoint ``fake clusters'' of $2{,}000$ spans each,
drawn uniformly from the full type-A span pool---these mix every real
operator in corpus-rate proportions, so their top $n$-grams concentrate
on high-frequency meta-phrases (``now we need'', ``let me think'') with
no operator-specific signal.

\paragraph{Statistical comparisons.}
We test whether the within-real distribution stochastically dominates
each of the two reference distributions using a one-sided Mann--Whitney
$U$ test (alternative $=$ greater). For interpretability we report the
rank-biserial effect size
$r_\mathrm{rb} = 2U / (n_1 n_2) - 1$,
which rescales $U$ to $[-1, +1]$ and equals $2\,P(A > B) - 1$: with
sample sizes in the thousands the $U$-test rejects for trivial gaps, so
$r_\mathrm{rb}$ is the load-bearing summary of \emph{how much} the
distributions are separated. The 435 within-cluster pair cosines per
cluster are not statistically independent---they share the 30 underlying
seed embeddings---which makes the $U$-test $p$-values anticonservative;
$r_\mathrm{rb}$ does not depend on this independence assumption and is
therefore reported as the primary result.

\begin{figure}[ht]
\centering
\includegraphics[width=0.72\linewidth]{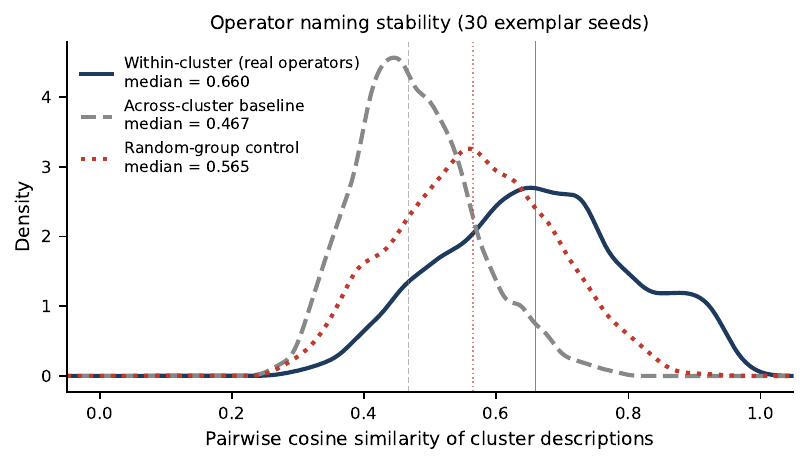}
\caption{Naming stability: pairwise cosine similarity distributions
between cluster-description embeddings generated under 30 exemplar
seeds (\texttt{text-embedding-3-small}, $L_2$-normalized). The
\emph{within-cluster} distribution (median $= 0.660$) is well
separated from both the \emph{across-cluster} baseline (median $=
0.467$) and the \emph{random-group control} (median $= 0.565$), which
applies the same naming protocol to seven disjoint random span
groupings. The gap confirms that operator names are stable at the
level of meaning and are not an artifact of the LLM prompt.}
\label{fig:stability}
\end{figure}

\paragraph{Results and interpretation.}
Figure~\ref{fig:stability} shows the three cosine-similarity distributions.
The within-real pooled median cosine is $0.660$ versus $0.467$ for the
across-real baseline and $0.565$ for the random-group control.
Real-cluster names dominate the across-cluster baseline with
$r_\mathrm{rb} = +0.72$ and the random-group control with
$r_\mathrm{rb} = +0.39$ (one-sided $U$-test $p \approx 0$ and
$p < 10^{-149}$, respectively): roughly $86\%$ of within-real pairs
are more similar than a random across-cluster pair, and $69\%$ more
similar than a random within-fake-group pair. Every individual operator
clears the across-cluster baseline (per-cluster $U$-test $p < 10^{-12}$;
$r_\mathrm{rb} \in [+0.22\ (\textsc{grounding}, \mathrm{weakest}),\
+1.00\ (\textsc{backtracking}, \mathrm{strongest})]$). The random-group
control falls midway between within-real and across-cluster
($r_\mathrm{rb} = +0.44$ of random vs.\ across; median gap $= 0.098$):
LLM descriptions of mixed-operator span pools are somewhat self-consistent,
but far below the within-real ceiling, confirming that the coherence of
real operator names is content-driven rather than a generic tendency of
the naming prompt to produce similar-sounding sentences.

\section{Additional Analyses}
\label{app:additional}

\subsection{Transition Structure and Run-Length Statistics}

\begin{figure}[t]
\centering
\includegraphics[width=\linewidth]{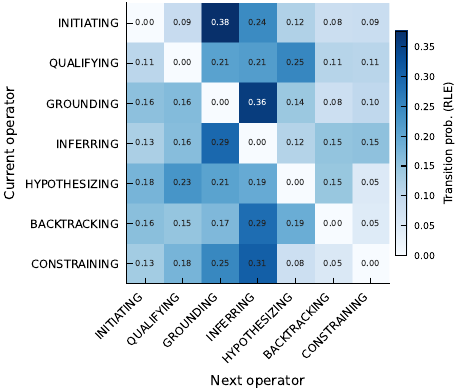}
\caption{Left: operator transition probability matrix. Entry $(i,j)$ = probability that operator $j$ follows $i$. Right: run-length statistics per operator---mean and distribution of consecutive same-operator span counts. \textsc{Grounding} has the longest self-runs (sustained fact-anchoring); \textsc{Backtracking} is nearly always a single span.}
\label{fig:transition_rle}
\end{figure}

Figure~\ref{fig:transition_rle} shows the operator transition probability matrix and run-length statistics. \grounding has the longest self-runs, reflecting sustained fact-anchoring that persists for multiple consecutive spans. \backtracking is nearly always a single span---it interrupts but does not persist---which is consistent with its function as a brief error-signal rather than a sustained strategy change. The \hypothesizing $\to$ \inferring pathway has elevated probability relative to most other off-diagonal entries, consistent with a generate-and-test schema in which tentative scenarios are followed by conclusions.

\subsection{Model Identification Confusion Matrix}
\label{app:model_id_confusion}

\begin{figure}[t]
\centering
\includegraphics[width=\linewidth]{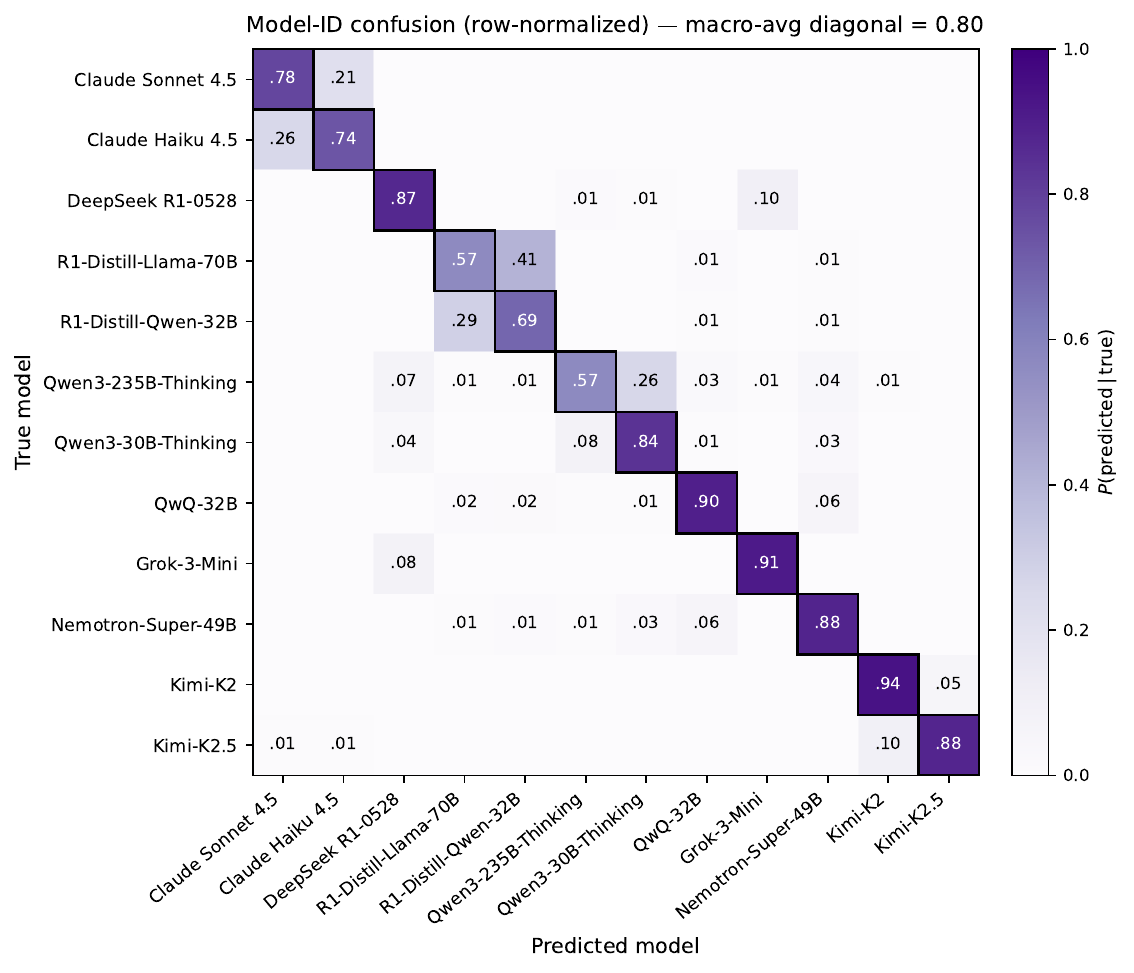}
\caption{Model identification confusion matrix (row-normalized) under problem-level 5-fold CV with the full Op-XGB feature recipe (Appendix~\ref{app:opxgb}). Accuracy = $0.799$, macro-AUC = $0.987$. Errors concentrate within model families (R1-Distill pair, Claude pair, Kimi pair).}
\label{fig:model_id_confusion}
\end{figure}

Figure~\ref{fig:model_id_confusion} shows the confusion matrix for the 12-way model identification task under problem-level 5-fold CV using the full Op-XGB feature recipe (Appendix~\ref{app:opxgb}). The classifier reaches accuracy $= 79.9\%$ and macro-AUC $= 0.987$. Errors concentrate within model families. The R1-Distill-Llama-70B / R1-Distill-Qwen-32B pair contributes the largest off-diagonal mass: R1-Distill-Llama-70B traces are predicted as R1-Distill-Qwen-32B at $0.41$ and the reverse at $0.29$, consistent with their shared distillation lineage from DeepSeek-R1. The Claude-Haiku-4.5 / Claude-Sonnet-4.5 pair shows mutual confusion in the $0.21$--$0.26$ range, and Kimi-K2.5 is predicted as Kimi-K2 at $0.10$ (with the reverse at $0.05$). Off-diagonal mass outside these family clusters is small. Per-model AUCs span $0.967$ (R1-Distill-Llama-70B) to $0.997$ (Kimi-K2, Kimi-K2.5, Grok-3-Mini, all tied at this rounding).

\clearpage

\bibliographystyleS{unsrt}
\bibliographyS{appendix_citations}

\end{document}